\newcommand{\cadre}[1]{%
\begin{center}
\noindent\fbox{\parbox[b]{.95\linewidth}{#1}}
\end{center}}%

\documentstyle[aclap]{article}

\title{\vspace{-0.5in}Coordination as a Direct Process}
\author{Augusta Mela \\
LIPN-CNRS URA 1507 \\Universit\'e de Paris XIII \\ 93 430 Villetaneuse
FRANCE
\\{\tt am@ura1507.univ-paris13.fr}\And
Christophe Fouquer\'e \\ LIPN-CNRS URA 1507
 \\ Universit\'e de Paris XIII \\ 93 430 Villetaneuse FRANCE \\{\tt
cf@ura1507.univ-paris13.fr}}

\begin{document}
\bibliographystyle{fullname}
\maketitle
\vspace{-0.5in}

\begin{abstract}
We propose a treatment of coordination based on the concepts of functor,
argument and subcategorization.
Its formalization comprises two parts which are conceptually
independent. On one hand,
we have extended the feature structure unification to disjunctive and
set values in order to check
the compatibility and the satisfiability of subcategorization
requirements by structured
complements. On the other hand, we have considered the conjunction {\em
et (and)} as the head of the
coordinate structure, so that coordinate structures stem simply from the
subcategorization
specifications of {\em et} and the general schemata of a head
saturation.
Both parts have been encoded within HPSG using the same resource that is
the
subcategorization and its principle which we have just extended.
\end{abstract}

\section{Introduction}
Coordination has always been a centre of academic interest, be it in
linguistic theory or in
computational linguistics. The problem is that the assumption according
to only the constituents
of the same category (1) may be conjoined is false; indeed,
coordinations of different categories
(2)-(3) and of more than one constituent (4)-(5) should not be dismissed
though being marginal
in written texts and must
be accounted for\footnote{This research has been done for the French
coordination {\em et (and)}.}.

\vspace{0.2cm}\noindent
\begin{tabular}{p{.04\linewidth}p{.88\linewidth}}
(1) & Jean danse \underline{la valse} et \underline{le tango} \\
        &       {\footnotesize(Jean dances the waltz and the tango.)} \\
(2) &   Je sais \underline{son \^age} et \underline{qu'elle est venue
ici}. \\
        &       {\footnotesize(I know her age and that she came
here.)}\\
(3)&    Un livre \underline{int\'eressant} et \underline{que j'aurai du
plaisir}  \underline{\`a lire}.\\
        &       {\footnotesize(An interesting book and which I will
enjoy to read.)}\\
(4)&    Je demande \underline{\`a Pierre son v\'elo} et \underline{\`a
Marie} \underline{sa canne \`a p\^eche}.\\
        &       {\footnotesize(I ask Peter for his bike and Mary for her
fishing rod.)}\\
(5)&    Pierre \underline{vend un v\'elo} et \underline{donne une canne}
\underline{\`a p\^eche} \`a Marie.\\
        &               {\footnotesize(Peter sells a bike and gives a
fishing rod to Mary.)}
\end{tabular}\vspace{0.2cm}

We claim here that the ``local combinatory potential'' of lexical heads,
encoded in the
subcategorization feature, explains the previous linguistic facts:
conjuncts may be of different
categories as well as of more than one constituent, they just have to
satisfy the
subcategorization constraints.\par
We focus here on the coordination of syntagmatic categories (as opposite
of lexical categories).
More precisely, we account for cases of non constituent coordination
(4), of Right
Node Raising  in coordination (5) but not for cases of Gapping.\par

Our approach which is independent of any framework, is easily and
precisely encoded in
the formalism of Head Driven Phrase Structure Grammar (HPSG)
\cite{Pollard94}, which is
based on the notion of head and makes available the feature sharing
mechanism we need.
The paper is organized as follows. Section 2 gives a brief description
of basic
data and discusses some constraints and available structures. Section 3
summarizes previous approaches
and section 4 is devoted to our approach. The french coordination with
{\em et} serves throughout the
paper as an example.

\section{A brief description of Basic Data and Constraints}
The classical typology of coordination, i.e. coordination of
constituents (1) and
of non-constituents, hides some regularity of the phenomenon as it
focuses
on concepts of constituent and syntactic category.

A coordination of constituents is interpreted as one phrase without any
gap. The constituents may be of the same category (1) as well as of
different categories (2)-(3). However, this last case is constrained as
examplified hereafter\footnote{The star * marks ungrammatical
sentences.}.

\vspace{0.2cm}\noindent
\begin{tabular}{p{.04\linewidth}p{.88\linewidth}}
(2) &   Je sais \underline{son \^age} et \underline{qu'elle est venue
ici}. \\
        &       {\footnotesize(I know her age and that she came
here.)}\\
(2a)&   Je sais \underline{son \^age} et \underline{son adresse}. \\
        &       {\footnotesize(I know her age and her address.)}\\
(2b)&   Je sais \underline{qu'elle a 30 ans} et \underline{qu'elle est
venue ici}. \\
        &       {\footnotesize(I know that she is 30 and that she came
here.)}\\
(2c)&   *Je sais \underline{\`a Marie} et \underline{qu'elle est venue
ici}. \\
        &       *{\footnotesize(I know to Marie and that she came
here.)}\\
(2d)&   Je demande \underline{l'addition} et \underline{que quelqu'un
paie}. \\
        &       {\footnotesize(I ask for the bill and for someone to
pay.)}\\
(2e)&   *Je rends \underline{l'addition} et \underline{que quelqu'un
paie}. \\
        &       *{\footnotesize(I give back the bill and someone to
pay.)}\\
\end{tabular}\vspace{0.2cm}
In these examples, the coordinate structure acts as the argument of the
verb.
This verb must subcategorize for each constituent of the coordination
and this
is not the case in example (2c)-(2e). Note that modelizing coordination
of different
categories as the unification (i.e. underspecification) of the different
categories
would lead to
accept the six examples or wrongly reject (2d) according to the
descriptions
used\footnote{Apart from {\em ad hoc} modelizations.}.

Coordination of more than one constituent are often classified as
Conjunction Reduction (4), Gapping (1a-1b) and Right Node Raising (5)
\cite{Hudson76}.

\vspace{0.2cm}\noindent
\begin{tabular}{p{.04\linewidth}p{.88\linewidth}}
(1a)&   Jean danse la valse et Pierre, le tango.\\
        &       {\footnotesize (Jean dances the waltz and Pierre the
tango.)}\\
(1b)&   Hier, Jean a dans\'e la valse et aujourd'hui, le tango.\\
        &       {\footnotesize (Yesterday, Jean danced the waltz and
today, the tango.)}
\end{tabular}\vspace{0.2cm}
In the case of Gapping structures\footnote{At least in SVO languages
written texts.}, the subject (1a) and/or an extracted element (1b)
is present in the two sides. The only allowed coordinated structure
is {\em [Jean danse la valse] et [Pierre le tango]} for (1a) and {\em
[Hier,
 Jean a dans\'e la valse] et [aujourd'hui, le tango]} for (1b)  as
wh-sentences on other
parts ({\em [la valse] et [Pierre]} or {\em [la valse] et [Pierre le
tango]}) are impossible.

A contrario, in the case of Conjunction Reductions, wh-sentences (4a,b)
as well as
cliticization (4c,d) are allowed referring to what follows the verb (as
for
coordination of constituents) and treating the arguments simultaneously
on
the two parts of the coordination:

\vspace{0.2cm}\noindent
\begin{tabular}{p{.04\linewidth}p{.88\linewidth}}
(4a)&   Je sais \`a qui demander un v\'elo et une canne \`a p\^eche.\\
        &       {\footnotesize(I know who I ask for a bike and for a
fishing rod.)}\\
(4b)&   Je sais \`a qui les demander.\\
        &       {\footnotesize(I know who I ask for them.)}\\
(4c)&   Je leur demande un v\'elo et une canne \`a p\^eche.\\
        &       {\footnotesize(I ask them for a bike and for a fishing
rod.)}\\
(4d)&   Je les leur demande.\\
        &       {\footnotesize(I ask them for them.)}\\
\end{tabular}\vspace{0.2cm}

Let us remark that a comma is inserted between {\em \`a Marie} and {\em
sa canne
\`a p\^eche} in case of extraction before {\em et} as in (1b),
indicating the two
sentences have not necessarily to be analyzed in the same way:

\vspace{0.2cm}\noindent
\begin{tabular}{p{.04\linewidth}p{.88\linewidth}}
(4e)&   Je demande \`a Pierre son v\'elo et \`a Marie sa canne \`a
p\^eche.\\
        &       {\footnotesize(I ask Peter for his bike and Marie for
her fishing rod.)}\\
(4f)&   A Pierre, je demande son v\'elo et \`a Marie, sa canne \`a
p\^eche.\\
        &       {\footnotesize(Peter, I ask for a bike and Marie, for a
fishing rod.)}\\
\end{tabular}\vspace{0.2cm}

Two structures are available in case of Conjunction Reductions. One
structure
corresponds to a coordination of sentences with a gap of the verb after
{\em et},
the other one consists in taking the coordinate parallel sequence of
constituents as only one structure. The previous facts argue for the
second
possibility (see also section 3 for criticism of deletion approach).
\par
Last, note that gapping the verb is less compatible with head-driven
mechanisms (and the comma in (4f) could be such a head mark, see
\cite{BEF96} for
an analysis of Gapping coordinations). It seems then that
the structure needed for Conjunction Reduction is some generalization of
the standard structure used for coordination of constituents. Our
proposal
is then focused on this extension. We do not care of Gapping cases as
their
linguistic properties seem to be different.

It remains to integrate Right-Node Raising and to extend these cases to
more
complicated ones. Section 4 includes examples of such cases and shows
that
our proposal can manage them adequately.

\section{Previous Approaches}
There exists a classical way to eschew the question ``what can be
coordinated ?'' if one
assumes a deletion analysis. Indeed, according to this approach
\cite{Chomsky57,Banfield81},
 only coordination of sentences are basic and other syntagmatic
coordinations should be
considered as coordinations of reduced sentences, the reduction being
performed by deleting
repeated elements. This approach comes up against insurmountable
obstacles, chiefly with the problem of applying transformation in
reverse, in the analysis process \cite{Schachter73}.\par
A direct approach has been proposed at once by
Sag \& al. \cite{Sag85} within the framework of Generalized Phrase
Structure
Grammar (GPSG), by \cite{Pollard94} within HPSG, and \cite{Bresnan86}
within
Lexical Functional Grammar (LFG). These approaches have tried to account
for
coordination of different categories in reducing the constraint from
requiring the same category for conjuncts to a weaker constraint of
category compatibility. Whatever the nature of subcategorization
information may
be, syntactical in GPSG, hybrid in HPSG, functional in LFG, two
categories
are compatible if they subsume a ``common denominator'', in this case a
common partial structure.

Technically, the compatibility is checked by computing a
``generalization'' of categories and imposing the generalization
comprises all features expected in
the given context. For example, the context in (6), that is, the verb
{\em \^etre (to be)},
expects a
predicative argument and both categories NP and AP are just predicative
categories.

\vspace{0.2cm}\noindent
\begin{tabular}{p{.04\linewidth}p{.88\linewidth}}
(6)&    Il est \underline{le p\`ere de Marie} et \underline{fier de
l'\^etre}.\\
        &       {\footnotesize(He is Mary's father and proud of it.)}
\end{tabular}\vspace{0.2cm}
 
However, this solution cannot be applied generally because all
coordinations have not
such ``natural'' intersection (see (2)). So we claim that we have
nothing else to do but explicitly
enumerate, within the head subcategorization feature, all the structures
allowed as complement.

\section{Our Approach}
Our proposition involves three stages. We begin by formulating
constraints on coordinate
structures, then we define how to build the coordinate structures and we
end by specifying how
the previous constraints filter through such coordinate structures.

\subsection{Constraints on coordinate structures}
In order to precisely formulate the constraints on coordinate
structures, we distinguish
the role of functor and that of argument, where
functor categories are those that bear unsatisfied subcategorization
requirements,
as it is the case in Categorial Grammars \cite{Dowty88,Steedman90}.
Lexical heads
(1) are functors in relation to the arguments they select and, by
composition, any
expression that contains an unsaturated functor is a functor (5)-(7).

\vspace{0.2cm}\noindent
\begin{tabular}{p{.04\linewidth}p{.88\linewidth}}
(7)&    Il \underline{pr\'etend d\'etester} et refuse ces beaux spots
lumineux.\\
        &       {\footnotesize(He claims to hate and refuses these
beautiful spotlights.)}
\end{tabular}\vspace{0.2cm}

Arguments are the complements selected by the head\footnote{In this
paper, we
restrict arguments to complements. In our HPSG encoding, they are
treated in
the SUBCAT feature. In a Borsley-like manner, we suppose a special
feature
for the subject. However, our approach can be generalized to subjects.}.
An
argument may often be realized
by different categories.
For example, the argument required by {\em savoir (to know)} may be a NP
or a Completive: we say that the requirement is disjunctive and we
represent
the different
alternatives within subcategorization feature disjunctive values. An
argument
specification is then a disjunction of categories. When the lexical head
requires
several complements ({\em to ask somebody something}), the requirement
is said
multiple or n-requirement. To the extent that disjunction only appears
in argument
specifications, a n-requirement is a multi-set of simple requirements.
The
choice of set (or more precisely multiset) rather than list value for
the
feature SUBCAT allows us to account
for  {\em Je demande \`a Pierre  son v\'elo}  as well as {\em Je demande
son
v\'elo \`a Pierre}. Gunji \cite{Gunji87} makes the same choice.
However our criterion can be formalized in a theory whose order of
arguments
obeys to an obliqueness hierarchy.

\vspace{0.5cm}
{\bf Requirement inheritance}.
A functor may compose with another functor or with arguments. In
functor-arguments
composition, the resulting expression inherits the unsatisfied
requirement from the functor
when it is not empty. For example, in (5), both conjuncts inherit the
unsatisfied requirement
from their heads.  Likewise the functor composition inherits a
requirement from the unsatisfied
functor\footnote{In functor composition, functors cannot be both
unsaturated:
{\em * Il \underline{promet de manger} \`a sa m\`ere des bananes.(* he
promises to eat his mother bananas.)}, cf. the Incomplete Constituent
Constraint
\cite{Pollard94}.}. In (7), {\em pr\'etend d\'etester} inherits the
unsatisfied
requirement of {\em d\'etester}, i.e. the
requirement of an object.

\vspace{0.5cm}
{\bf Adjuncts}.
To account for the continuum which exists from strictly subcategorized
complements to
adjuncts, we adopt the hypothesis suggested by \cite{Miller91} according
to which adjuncts could be
accorded the same status as arguments by integrating them into the
subcategorization
requirement through an optional lexical rule. That would enable us to
account for coordination
of adjuncts of different categories (3) as well as coordination of more
than one constituent with
adjuncts (10)-(11) below. \par
Let us remark that syntactically the inserted modifiers behave as
complements but semantically they still can behave as modifiers, i.e.
the semantic features (CONTENT in HPSG) of the resulting
Head-Complements structure is token-identical to that of the adjunct.
See \cite{AAGDb96} for such an analyze of adverbs as complements.\par
 Last, note that these lexical rules can be interpreted statically
as well as dynamically. In the first case, the extended lexicon is
pre-computed
and requires no runtime application.

\vspace{0.5cm}
{\bf Satisfiability conditions of requirements}.
 We observe here that a coordination of different categories may appear
as head
complement when the head requirement is disjunctive and a coordination
of more than one
constituent appears when such a requirement is multiple. Last, functors
may conjoin when their
subcategorization requirements are compatible. These observations are
synthesized in one
coordination criterion.

The first observation is summarized in (C1) and illustrated in (2').\par

\cadre{(C1) A subcategorization 1-requirement is satisfied either by one
of
the disjuncts or by a coordination of disjuncts.} \par

\vspace{0.2cm}\noindent
\begin{tabular}{p{.04\linewidth}p{.88\linewidth}}
(2')&   Je sais son \^age /qu'elle est venue ici / son \^age et qu'elle
est venue ici.\\
        &       {\footnotesize(I know her age /that she came here / her
age and that she came here.)}
\end{tabular}\vspace{0.2cm}

The second one is illustrated below, where subcategorization
n-requirements are satisfied
either by:
\begin{itemize}
\item a series of n complements which satisfy respectively the n
requirements

\noindent
\begin{tabular}{p{.04\linewidth}p{.88\linewidth}}
(8)&    Je demande \underline{\`a Pierre} \underline{son v\'elo et sa
canne} \underline{\`a p\^eche}.\\
        &       {\footnotesize(I ask Peter for his bike and for his
fishing rod.)}
\end{tabular}
\item a coordination of a series of this kind

\noindent
\begin{tabular}{p{.04\linewidth}p{.88\linewidth}}
(9)&    Je demande \underline{\`a Pierre} \underline{son v\'elo} et
\underline{\`a Marie} \underline{d'o\`u elle vient}.\\
        &       {\footnotesize(I ask Peter for his bike and Mary where
she comes from.)}
\end{tabular}
\item a coordination may concern sub-series of arguments

\noindent
\begin{tabular}{p{.04\linewidth}p{.88\linewidth}}
(10)&   Pierre a achet\'e \underline{un livre \`a Marie} et
\underline{un disque \`a Pierre} pour 100F.\\
        &       {\footnotesize(Peter has bought a book for Mary and a CD
for Peter for 20\$.)}
\end{tabular}
\item or sequences of more than one constituent with adjuncts

\noindent
\begin{tabular}{p{.04\linewidth}p{.88\linewidth}}
(11)&   J'ai vu \underline{Pierre hier} et \underline{Marie lundi}.\\
        &       {\footnotesize(I have seen Peter yesterday and Mary
monday.)}
\end{tabular}
\item or adjuncts of different categories

\noindent
\begin{tabular}{p{.04\linewidth}p{.88\linewidth}}
(3)&    Un livre \underline{int\'eressant} et \underline{que j'aurai du}
\underline{plaisir \`a lire}.\\
        &       {\footnotesize(An interesting book and which I will
enjoy to read.)}\\
\end{tabular}
\end{itemize}
All these situations are summarized in (C2): \par

\cadre{(C2) A subcategorization n-requirement is satisfied by m
arguments,$0<m\leq n$,
either by a sequence of m arguments such that each argument satisfies
one and only one
element of the requirement or by a coordination of such sequences.
The result has a $n-m$ requirement.} \par

Note that (C1) and (C2) should be computed simultaneously in order to
account for
structures as (9). The notion of partial saturation in (C2) allows us to
account for coordination
of sub-series of arguments as in (10).

\vspace{0.5cm}
{\bf Functors coordination and compatibility of requirements}.
Functors may be simple (1), composed (7), of different structures (12)
or partially
saturated (13)-(5).

\vspace{0.2cm}\noindent
\begin{tabular}{p{.04\linewidth}p{.88\linewidth}}
(12)&   Je pense \underline{offrir} et \underline{que je recevrai} des
cadeaux. \\
        &       {\footnotesize(I think to offer and that I will receive
gifts.)} \\
(13)&   Je pense \underline{recevoir de Jean} et \underline{offrir \`a
Pierre} du caviar de Russie.\\
        &       {\footnotesize(I expect to receive from John and offer
to Peter Russian caviar.)}
\end{tabular}\vspace{0.2cm}

In all cases, when they are conjoined, they share their arguments: there
must therefore
exist at least one possibility of satisfying them simultaneously. In
this case, the unification of
their subcategorization requirements succeeds and they are said to be
compatible and the two
functors may be conjoined. This unification has to account for
disjunctive values.

\vspace{0.5cm}
{\bf Coordination criterion : satisfying and imposing requirements}.
As an entity can be both functor and argument (12)-(13) our coordination
criterion
(necessary condition) is the following one: the conjuncts must satisfy
the same
simple or multiple subcategorization
requirement and impose compatible subcategorization requirements.

\subsection{Computing the subcategorization requirements compatibility}
We have now to define an extension of the usual unification $\cup$ of
structures in order to
compute the subcategorization requirements compatibility. This extension
is an internal
operation over the subcategorization requirements which accounts for
disjunctive and set
values. ${{\cal U}}$ is the unification of argument specifications
defined from
$\cup$, ${{\cal U}}^+$ is its extension
to n-requirements.

\begin{itemize}
\item {\bf Unification of two argument specifications $\alpha$ and
$\beta$}.

Let us have
        $\alpha = \bigvee_{k=1\dots p} s_k, \beta = \bigvee_{l=1\dots q}
t_l$, with categories $s_k, t_l$, then

{\setlength\tabcolsep{0pt}\begin{tabular}{ll}
$\alpha {{\cal U}} \beta =$ & $\bigvee_{k,l} s_k \cup t_l$ for $k,l$
s.t. $s_k \cup t_l$ exists\\
                        & undefined if $s_k \cup t_l$ does not exist,
$\forall k,l$
\end{tabular}}

\item {\bf Unification of two n-requirements $\Phi$ and $\Psi$}.
$\Phi =\{\alpha_i / i\in[1,n] \}$ and $\Psi =\{\beta_i / i\in [1,n]\}$
be 2 n-requirements, where $\alpha_i$ and $\beta_i$ are
argument specifications,
the extended unification ${{\cal U}}^+$ of $\Phi$ and $\Psi$ is defined
if there exists a permutation $p$ on
$[1,n]$ such that $\alpha_i {{\cal U}} \beta_{p[i]}$  exists $\forall i
\in[1,n]$.
In this case $\Phi {{\cal U}}^+ \Psi = \{\alpha_i {{\cal U}}
\beta_{p[i]} / i \in [1,n]\}$ else  $\Phi {\cal U}^+ \Psi$ is undefined.
\end{itemize}

\cadre{Two n-requirements are compatible iff their unification ${\cal
U}^+$ succeeds.}

We consider that conjoined functors should have the same
valence\footnote{This
condition will forbid the conjunction of e.g. verbs with SUBCAT lists of
different lengths, but which
would have a unification under the alternative interpretation, thus
avoiding
sentences like {\em *John bought and gave the
book to Mary}, \cite{Miller91}.}.
Note that the unification of two n-requirements is ambiguous because we
may
have several permutations which lead to success.

\subsection{How coordinate structures are built}
Until now we have just defined constraints on the coordinate structures
but we did not
mention how these structures are built. We want that a coordinate
structure inherits features
from its conjuncts without necessarily failing in case of conflicting
values. The generalization method
\cite{Sag85} has this objective but overgenerates because the
conflicting values are
ignored. In contrast, the use of composite categories \cite{Cooper91}
keeps
conflicting values within
the connective ``$\wedge$''. Intuitively, if {\em son \^age (her age)}
is a
NP and {\em qu'elle est venue ici (that she
came here)} is a Completive, {\em son \^age et qu'elle est venue ici
(her
age and that she came here)} is a
conjunctive composite category NP$\wedge$Compl.

\vspace{0.5cm}
{\bf The structuring of categories : composite and tuple of categories}.
We propose to extend the operation $\wedge$ to complex categories and to
use a new
connective $<\dots>$ in order to define tuple of categories. With these
two
connectives, a total structuring of categories is possible and all the
coordinate structures may
have a status. For example, the underlined expression in (14) will be
represented by the
structured category:
\renewcommand{\arraystretch}{0.3}
\arraycolsep=0.5pt
$\left\langle
{\scriptstyle PP,}\left\lbrack
\begin{array}{l}
{\scriptstyle NP \wedge Compl} \\
{\scriptstyle Subcat ~PP}
\end{array}
\right\rbrack
\right\rangle$.

\vspace{0.2cm}\noindent
\begin{tabular}{p{.04\linewidth}p{.88\linewidth}}
(14)&   Je recommande \underline{\`a Pierre} \underline{la lecture} et
\underline{qu'il s'inspire} de la Bible.\\
&       {\footnotesize(I recommend to Peter the lecture and that he
inspires himself of the Bible.)}
\end{tabular}\vspace{0.2cm}

The extension to complex categories is not uniform. Coordinate structure
features are not
necessarily composites or tuples of corresponding features from each
conjunct. In fact,
features which are allowed to have conflicting values will be
compounded, whereas other
features as SUBCAT must unify.
This structuring is encoded later within the definition of the lexical
entry of {\em et}.
 
\vspace{0.5cm}
{\bf Lexicalization of the coordination rule}.
We consider, as in \cite{Paritong92}, the conjunction {\em et} as the
head of the coordinate structure.
Consequently, coordinate structures no longer have to be postulated in
the grammar by a special
rule of coordination: they stem simply from the general schemata of the
head saturation and the
subcategorization specifications of the conjunction. For sake of
simplicity,
only binary coordination is treated here. \cite{Paritong92} accounts for
multiple coordination as a binary structure where the comma has a
similar
function as a lexical conjunction. With that one restriction, the
HPSG-like
lexical entry of {\em et} can be:

\begin{center}
$\left\lbrack
\begin{array}{l}
{\scriptstyle Phon ~\setminus et\setminus} \\
{\scriptstyle Synsem {\scriptscriptstyle
~<[1],\dots,[M]>\wedge<[1'],\dots,[M']>}|Cat=} \\
~~~\left\lbrack
        \begin{array}{l}
        {\scriptstyle Part ~<C_1,\dots,C_M>\wedge <C'_1,\dots,C'_M>} \\
        {\scriptstyle Subcat~\left\lbrace
                \begin{array}{l}
                {\scriptstyle [1]}\left\lbrack
                \begin{array}{l}
                {\scriptstyle Part ~C_1} \\
                {\scriptstyle Subcat ~\{\}}
                \end{array}
                \right\rbrack,
                {\scriptstyle \dots,}
                {\scriptstyle [M]}\left\lbrack
                \begin{array}{l}
                {\scriptstyle Part~ C_M} \\
                {\scriptstyle Subcat ~\Phi_M}
                \end{array}
                \right\rbrack,\\
                {\scriptstyle [1']}\left\lbrack
                \begin{array}{l}
                {\scriptstyle Part ~C_1'} \\
                {\scriptstyle Subcat ~\{\}}
                \end{array}
                \right\rbrack,
                {\scriptstyle \dots,}
                {\scriptstyle [M']}\left\lbrack
                \begin{array}{l}
                {\scriptstyle Part~ C_M'} \\
                {\scriptstyle Subcat ~\Phi_M'}
                \end{array}
                \right\rbrack,\\
                {\scriptstyle \Phi_M {\cal U}^+ \Phi_M'}
                \end{array}
        \right\rbrace}
        \end{array}
        \right\rbrack
\end{array}
\right\rbrack$
\end{center}

The following LP-constraint on the lexical entry of {\em et} ensures the
correct order of
conjunction and conjuncts:
\begin{center}
$[i] < conj <  [i']$, where $i \in [1,M], i' \in [1',M']$.
\end{center}
This LP-constraint is the minimum required to distinguish the two parts
of the coordinate structure. However, the functor this coordinate
structure
(partially-)saturates may impose its own LP-constraint (e.g. an
obliqueness hierarchy). In such a case, this LP-constraint has to be
satisfied
simultaneously by the two sets $\{[1],\dots,[M]\}$ and
$\{[1'],\dots,[M']\}$.

To represent the inheritance of the complements, here $\Phi_M {\cal U}^+
\Phi_M'$, we use a mechanism
of argument composition inspired by \cite{Hinrichs94}: the conjunction
{\em et} takes as
complements the two conjuncts $<C_1,\dots,C_M>$ and $<C'_1,\dots,C'_M>$
which may remain unsaturated
for their complements $\Phi_M$ and $\Phi_M'$, and the set $\Phi_M {\cal
U}^+ \Phi_M'$.
The coordination of m-tuples, as well as the coordination of simple
conjuncts ($M=1$)
stems from the saturation of the conjunction {\em et}. As noted in
4.1., only the last element of the
tuple $C_M$ (or $C'_M$) can be unsaturated and be the source of
inheritance.  Example of resulting HPSG-like
analysis is given in figure~\ref{analyse1} for the underlined phrase in
(15).

\vspace{0.2cm}\noindent
\begin{tabular}{p{.04\linewidth}p{.88\linewidth}}
(15)&   Jean conseille \underline{\`a son p\`ere d'acheter et \`a sa}
\underline{m\`ere d'utiliser}  un lave-vaisselle.\\
        & {\footnotesize(Jean advises his father to buy and his mother
to use a dish washer.)}
\end{tabular}\vspace{0.2cm}

\newcommand{\A}{{\footnotesize
$\left\lbrack
\begin{array}{l}
{\scriptstyle Phon ~\setminus \grave{a} ~son~ p\grave{e}re ~d'acheter
~et ~\grave{a} ~sa ~m\grave{e}re ~d'utiliser\setminus} \\
{\scriptstyle Synsem <[1],[2]>\wedge<[3],[4]>|Cat}\left\lbrack
        \begin{array}{l}
        {\scriptstyle Part ~<PP,Compl>\wedge <PP,Compl>} \\
        {\scriptstyle Subcat~\{NP\}}
        \end{array}
        \right\rbrack
\end{array}
\right\rbrack$
}}

\newcommand{\Ba}{{\footnotesize
$\left\lbrack
\begin{array}{l}
{\scriptstyle Phon ~\setminus \grave{a} ~son ~p\grave{e}re\setminus} \\
{\scriptstyle Synsem ~[1]|Cat[Part~PP]}
\end{array}
\right\rbrack$
}}

\newcommand{\Bb}{{\footnotesize
$\left\lbrack
\begin{array}{l}
{\scriptstyle Phon ~\setminus d'acheter\setminus} \\
{\scriptstyle Synsem [2]|Cat}\left\lbrack
        \begin{array}{l}
        {\scriptstyle Part~ Compl} \\
        {\scriptstyle Subcat~\{NP\}}
        \end{array}
        \right\rbrack
\end{array}
\right\rbrack$
}}

\newcommand{\Bc}{{\footnotesize
$\left\lbrack
\begin{array}{l}
{\scriptstyle Phon ~\setminus \grave{a} ~sa ~m\grave{e}re\setminus} \\
{\scriptstyle Synsem ~[3]|Cat[Part~PP]}
\end{array}
\right\rbrack$
}}

\newcommand{\Bd}{{\footnotesize
$\left\lbrack
\begin{array}{l}
{\scriptstyle Phon ~\setminus d'utiliser\setminus} \\
{\scriptstyle Synsem ~[4]|Cat}\left\lbrack
        \begin{array}{l}
        {\scriptstyle Part~ Compl} \\
        {\scriptstyle Subcat~\{NP\}}
        \end{array}
        \right\rbrack
\end{array}
\right\rbrack$
}}

\newcommand{\C}{{\footnotesize
$\left\lbrack
\begin{array}{l}
{\scriptstyle Phon ~\setminus et\setminus} \\
{\scriptstyle Synsem ~<[1],[2]>\wedge<[3],[4]>|Cat}\left\lbrack
        \begin{array}{l}
        {\scriptstyle Part ~<PP,Compl>\wedge <PP,Compl>} \\
        {\scriptstyle Subcat~\{[1]}\left\lbrack
                \begin{array}{l}
                {\scriptstyle Part ~PP} \\
                {\scriptstyle Subcat ~\{\}}
                \end{array}
                \right\rbrack,
                {\scriptstyle [2]}\left\lbrack
                \begin{array}{l}
                {\scriptstyle Part ~Compl} \\
                {\scriptstyle Subcat ~\{NP\}}
                \end{array}
                \right\rbrack,
                {\scriptstyle [3]}\left\lbrack
                \begin{array}{l}
                {\scriptstyle Part ~PP} \\
                {\scriptstyle Subcat ~\{\}}
                \end{array}
                \right\rbrack,
                {\scriptstyle [4]}\left\lbrack
                \begin{array}{l}
                {\scriptstyle Part ~Compl} \\
                {\scriptstyle Subcat ~\{NP\}}
                \end{array}
                \right\rbrack,
                {\scriptstyle NP}
                \}
        \end{array}
        \right\rbrack
\end{array}
\right\rbrack$
}}

\noindent\begin{figure*}[t]
\begin{center}
\noindent\arraycolsep=0.5pt
\unitlength=1cm
\renewcommand{\arraystretch}{0.3}
\begin{picture}(16.5,5)
\put(4,4){\A}
\put(0,2){\Ba}
\put(3.4,2){\Bb}
\put(8.3,2){\Bc}
\put(12,2){\Bd}
\put(0.1,0.1){\C}

\put(8,3.5){\line(-6,-1){5}}
\put(8,3.5){\line(-2,-1){2}}
\put(8,3.5){\line(3,-1){2}}
\put(8,3.5){\line(6,-1){5}}
\put(8,3.5){\line(0,-1){2.5}}
\end{picture}
\caption{Analysis of {\em \`a son p\`ere d'acheter et \`a sa m\`ere
d'utiliser}}
\label{analyse1}
\end{center}
\end{figure*}
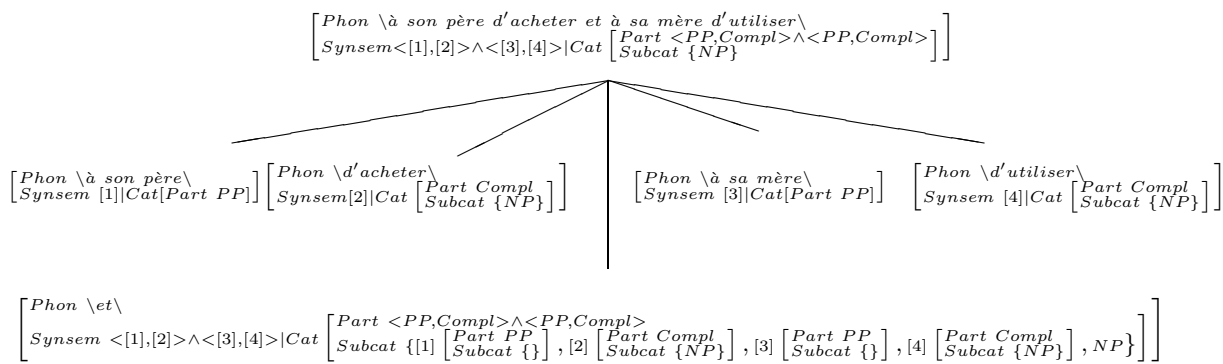

\subsection{How the constraints apply on coordinate structures}
We have now to define how arguments satisfy disjunctive and set
requirements.
Intuitively, if $\alpha_i$ is a (possibly disjunctive) argument
specification, an argument (possibly
composite) satisfies $\alpha_i$ iff each element of the composite
category matches one disjunct of
$\alpha_i$. Then, if $\Phi$ is a n-requirement, a tuple (or a
coordination
of tuples) of categories (possibly
composite) satisfies $\Phi$ iff each element of the tuple (for each
tuple)
satisfies one and only one
argument specification of $\Phi$.
More formally:

i) let $\alpha  = S^1 \vee \dots \vee S^{p}$ be an argument
specification,
and  $C = \bigwedge_{r=1\dots z} C_{r}$ be a composite category, then\\

\noindent
\cadre{\parbox[t]{.25\linewidth}{$C$ satisfies $\alpha$}
\parbox[b]{.05\linewidth}{iff}
\parbox[t]{.65\linewidth}{for each element of the composite category
$C$,there exists one disjunct of $\alpha$ that matches it}\\
\noindent
\parbox[t]{.23\linewidth}{~~~~~~~~~~~~~~~~~~~~~~~~~~~}
\parbox[b]{.05\linewidth}{(iff}
\parbox[t]{.65\linewidth}{$\forall r \in[1,z], \exists l \in[1,p] /
C_{r}  \cup S^l$ exists).}}

ii) let $\Phi$ be a n-requirement s.t.:

$\Phi =\{
\underbrace{S^1_1 \vee \dots \vee S^{p_1}_1}_{\alpha_1},
\dots ,
\underbrace{S^1_n \vee \dots \vee S^{p_n}_n}_{\alpha_n}
\}$

and $\Sigma$ be a coordination of $p$ tuples (if $p>1$) or one tuple (if
$p=1$) of composite categories $C^k_i$ s.t.:

$\Sigma = <C^1_1,\dots,C^1_n>\wedge \dots \wedge<C^p_1,\dots ,C^p_n>$

$C^k_i =\bigwedge_{r=1\dots z^k_i} C^k_{i,r}$

then\\

\noindent
\cadre{\parbox[t]{.25\linewidth}{$\Sigma$ satisfies $\Phi$}
\parbox[b]{.05\linewidth}{iff}
\parbox[t]{.65\linewidth}{each specification $\alpha_i$ has one and only
one
realization in each tuple of $\Sigma$} \\
\noindent
\parbox[t]{.23\linewidth}{~~~~~~~~~~~~~~~~~~~~~~~~~~~}
\parbox[b]{.05\linewidth}{(iff}
\parbox[t]{.65\linewidth}{$\forall k \in [1,p],\exists$ a permutation
$\pi_k$ on $[1,n]/
\forall i \in [1,n]~ C^k_{\pi_k[i]}$ satisfies $\alpha_i$).}}

Note that these requirement satisfiability conditions allows us to
account for examples
such as (9).

\subsection{A Coding in HPSG}
We extend
here the functor saturation schemata to the coordination case, within
the
framework of Head Driven Phrase Structure Grammar \cite{Pollard94}.

A subcategorization {\bf n-requirement} is satisfied by {\bf m
arguments}, $m\leq n$,
either by a sequence of m arguments (m-tuple) or by a coordination of
m-tuples.
The result has a $n-m$ requirement.

\vspace{0.5cm}
{\bf Saturation schemata}\footnote{$\Phi \cup \Psi$ is the set-union of
$\Phi$ and $\Psi$}
\begin{itemize}
\item[-] {\bf partial} ($\Psi \neq \{\}$) or total ($\Psi = \{\}$) of
saturated complements ($\Psi' = \{\}$)
\item[-] {\bf total} ($\Psi = \{\}$) of complements, the last being
partially  ($\Psi' \neq \{\}$)
                                                or totally saturated
($\Psi' = \{\}$)
\end{itemize}
\begin{center}
$\left\lbrack
\begin{array}{l}
Synsem|Cat[Subcat ~~\Psi\cup\Psi'] \\
Branches= \\
~~~~~~\left\lbrack
        \begin{array}{l}
        B-Head|Synsem|Cat[Subcat~~ \Phi\cup\Psi] \\
        B-Comp=\Sigma[Subcat~~ \Psi']
        \end{array}
        \right\rbrack
\end{array}
\right\rbrack$
\end{center}
where $\Sigma$ satisfies $\Phi$ and:
\begin{itemize}
\item $\Phi = \{<S^1_1 \vee \dots \vee S^{p_1}_1> , \dots , <S^1_m \vee
\dots \vee S^{p_m}_m >\}$  m-requirement, $\Psi$  $n-m$ requirement
\item $\Sigma = <C^1_1,\dots,C^1_m> \wedge \dots \wedge
<C^q_1,\dots,C^q_m>$  coordination of q m-tuples (if $q>1$) or one m-
        tuple (if $q=1$) of composite Synsem  $C^k_i =
\bigwedge_{r=1\dots z^k_i} C^k_{i,r}$
\item $\Psi$ or $\Psi'$ must be empty
\end{itemize}

Example of resulting analysis is given in figure~\ref{analyse2} for the
underlined phrase in (15):

\vspace{0.2cm}\noindent
\begin{tabular}{p{.04\linewidth}p{.88\linewidth}}
(15)&   Jean \underline{conseille \`a son p\`ere d'acheter et \`a sa}
\underline{m\`ere d'utiliser  un lave-vaisselle}.\\
        &{\footnotesize(Jean advises his father to buy and his mother to
use a dish washer.)}
\end{tabular}\vspace{0.2cm}

\newcommand{\W}{{\footnotesize
$\left\lbrack
\begin{array}{l}
{\scriptstyle Phon ~\setminus conseille ~\grave{a} ~son~ p\grave{e}re
~d'acheter ~et ~\grave{a} ~sa ~m\grave{e}re ~d'utiliser ~un
~lave-vaisselle\setminus} \\
{\scriptstyle Synsem ~[VP]}
\end{array}
\right\rbrack$
}}

\newcommand{\T}{{\footnotesize
$\left\lbrack
\begin{array}{l}
{\scriptstyle Phon ~\setminus conseille ~\grave{a} ~son~ p\grave{e}re
~d'acheter ~et ~\grave{a} ~sa ~m\grave{e}re ~d'utiliser\setminus} \\
{\scriptstyle Synsem ~|VP[Subcat ~\{NP\}]}
\end{array}
\right\rbrack$
}}

\newcommand{\U}{{\footnotesize
$\left\lbrack
\begin{array}{l}
{\scriptstyle Phon ~\setminus un ~lave-vaisselle\setminus} \\
{\scriptstyle Synsem ~[Part ~NP]}
\end{array}
\right\rbrack$
}}

\newcommand{\V}{{\footnotesize
$\left\lbrack
\begin{array}{l}
{\scriptstyle Phon ~\setminus conseille\setminus} \\
{\scriptstyle Synsem |Cat}\left\lbrack
        \begin{array}{l}
        {\scriptstyle Part ~V }\\
        {\scriptstyle Subcat~\{PP,NP\vee Compl\}}
        \end{array}
        \right\rbrack
\end{array}
\right\rbrack$
}}

\noindent\begin{figure*}[t]
\begin{center}
\unitlength=1cm
\renewcommand{\arraystretch}{0.3}
\arraycolsep=0.5pt
\noindent\begin{picture}(16.5,4)
\put(3,3.5){\W}
\put(1,2){\T}
\put(11,2){\U}
\put(0.1,0.1){\V}
\put(6,0.1){\A}

\put(7,3){\line(-3,-1){1.6}}
\put(7,3){\line(6,-1){3.6}}

\put(3,1.5){\line(-3,-1){1.6}}
\put(3,1.5){\line(6,-1){5}}

\end{picture}
\caption{Analysis of {\em conseille \`a son p\`ere d'acheter et \`a sa
m\`ere d'utiliser un lave-vaisselle}}
\label{analyse2}
\end{center}
\end{figure*}
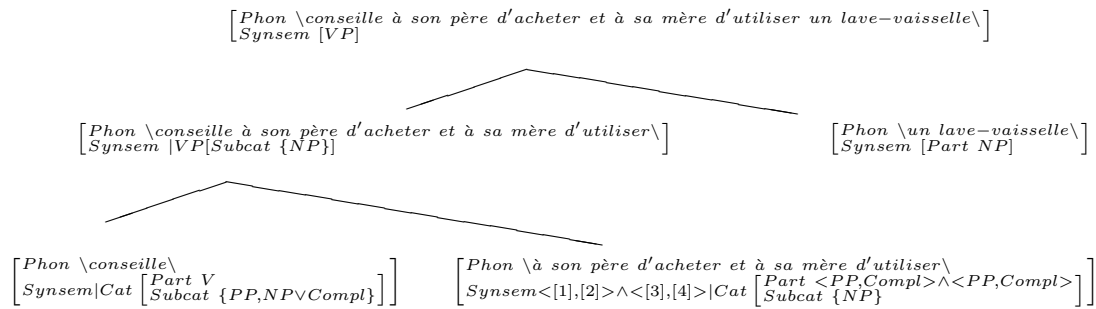

Note that the utility of partially saturated structures is empirically
supported by independent french data as argued for complementation of
tense auxiliaries  in \cite{AAGD94,AAGDa96}. \\ \par
Last, let us recall that within a theory as HPSG which integrates
syntactic and semantic information in
a single representation, a whole range of lexically determined
dependencies, e.g. case
assignment, government (of particular prepositions) and role assignment,
are modeled at the
same time via subcategorization because the value of subcategorization
feature is a complex of
syntactic and semantic information.

\section{Conclusion}
This approach based on concept of functor, argument and
subcategorization allows us to
account for many coordination data.
Its formalization comprises two parts which are conceptually
independent. On one hand,
we have extended the feature structure unification to disjunctive and
set values in order to check
the compatibility and the satisfiability of subcategorization
requirements by structured
complements. On the other hand, we have considered the conjunction {\em
et} as the head of the
coordinate structure, so that coordinate structures stem simply from the
subcategorization
specifications of {\em et} and a general schemata of the head
saturation.
Both parts have been encoded within HPSG using the same resource that is
the
subcategorization and its principle which we have just extended.

It remains to know in which extent our approach can be used for other
linguistic phenomena with symetrical sequences of more
than one constituent (comparative constructions (16), alternative
constructions):

\vspace{0.2cm}\noindent
\begin{tabular}{p{.04\linewidth}p{.88\linewidth}}
(16)&   Paul donne autant \underline{de couteaux aux filles} que
\underline{de pi\`eces aux gar\c cons}.\\
&       {\footnotesize(Paul gives as much knives to the girls as coins
to the boys.)}
\end{tabular}

\end{document}